\documentclass{bvm} %

\begin{document}

\newcommand{\bvmyear}{2026}

\selectlanguage{english} %

\title{ULS+\subtitle{Data-driven Model Adaptation Enhances Lesion Segmentation}}

\subtitle{}

\titlerunning{ULS+}

\author{
	\fname{Rianne} \lname[0009-0003-9601-0405]{Weber} \equalContribution \affiliation{Radboudumc} \authorsEmail{Rianne.Weber@radboudumc.nl} \street{Geert Grooteplein Zuid} \housenumber{10} \zipcode{6525 GA} \city{Nijmegen} \country{Netherlands} \isResponsibleAuthor,
	\fname{Niels} \lname[0009-0002-3072-4109]{Rocholl}\equalContribution \affiliation{Radboudumc} \authorsEmail{Niels.Rocholl@radboudumc.nl} \street{Geert Grooteplein Zuid} \housenumber{10} \zipcode{6525 GA} \city{Nijmegen} \country{Netherlands},
   \fname{Max} \lname{de Grauw} \affiliation{Radboudumc} \authorsEmail{Max.deGrauw@radboudumc.nl} \street{Geert Grooteplein Zuid} \housenumber{10} \zipcode{6525 GA} \city{Nijmegen} \country{Netherlands},
   \fname{Mathias} \lname[0000-0001-8157-8055]{Prokop} \affiliation{Radboudumc} \authorsEmail{Mathias.Prokop@radboudumc.nl} \street{Geert Grooteplein Zuid} \housenumber{10} \zipcode{6525 GA} \city{Nijmegen} \country{Netherlands},
   \fname{Ewoud} \lname[0000-0002-7090-2742]{Smit} \affiliation{Radboudumc} \authorsEmail{Ewoud.Smit@radboudumc.nl} \street{Geert Grooteplein Zuid} \housenumber{10} \zipcode{6525 GA} \city{Nijmegen} \country{Netherlands},
   \fname{Alessa} \lname[0000-0002-7602-803X]{Hering}\affiliation{Radboudumc} \authorsEmail{Alessa.Hering@radboudumc.nl} \street{Geert Grooteplein Zuid} \housenumber{10} \zipcode{6525 GA} \city{Nijmegen} \country{Netherlands}
}
\authorrunning{Weber et al.}

\institute{
Radboud university medical center
}

\email{Rianne.Weber@radboudumc.nl}

\maketitle

\begin{abstract}
    In this study, we present ULS+, an enhanced version of the Universal Lesion Segmentation (ULS) model. The original ULS model segments lesions across the whole body in CT scans given volumes of interest (VOIs) centered around a click-point. Since its release, several new public datasets have become available that can further improve model performance. ULS+ incorporates these additional datasets and uses smaller input image sizes, resulting in higher accuracy and faster inference.
    
    We compared ULS and ULS+ using the Dice score and robustness to click-point location on the ULS23 Challenge test data and a subset of the Longitudinal-CT dataset. In all comparisons, ULS+ significantly outperformed ULS. Additionally, ULS+ ranks first on the ULS23 Challenge test-phase leaderboard. By maintaining a cycle of data-driven updates and clinical validation, ULS+ establishes a foundation for robust and clinically relevant lesion segmentation models.\footnote{Code and weights available at \url{https://github.com/DIAGNijmegen/oncology-uls-plus}}
\end{abstract}

\section{Introduction}

Cancer remains one of the leading causes of mortality worldwide, and its burden is projected to continue rising. By 2050, the global cancer incidence is predicted to increase by 77\%, reaching 35.3 million cases~\cite{3886-Bizuayehu2024}. Correspondingly, radiologists are experiencing a steady growth in workload~\cite{3886-Bruls2020}. A substantial part of this workload arises from the longitudinal assessment of disease burden in oncologic imaging, where radiologists must identify, measure, and track target lesions over time according to the Response Evaluation Criteria in Solid Tumors (RECIST) guidelines. This process requires consistent lesion localization and quantification across time points, which is both time-consuming and prone to variability.

Artificial Intelligence (AI) offers opportunities to support radiologists in this workflow by automating parts of the lesion assessment process. AI-based methods have shown promise for lesion detection, segmentation, and classification, and can serve as building blocks for automated lesion tracking and response evaluation. Recent studies have demonstrated the feasibility of combining registration-based lesion matching with volumetric segmentation to support longitudinal tumor response assessment in CT imaging~\cite{3886-Hering2024}. These approaches aim to improve the consistency of lesion measurements and reduce observer variability in follow-up evaluations.

To further advance automatic lesion quantification, the Universal Lesion Segmentation (ULS) baseline model~\cite{3886-DeGrauw2025} was introduced in 2023 as part of the ULS23 challenge. ULS adopts a click-centered, interactive segmentation paradigm. Unlike methods designed for full-volume, exhaustive lesion segmentation, ULS allows a radiologist to select a specific lesion with a single click point. The model then rapidly returns a complete 3D mask from a localized Volume of Interest (VOI), thereby aligning with the focused, click-guided use case prevalent in interactive follow-up workflows.

Given the limited interaction required, as well as the lesion-specific training data used to develop this model, the ULS baseline has significant potential for use in clinical practice. However, since the release of the ULS model, new and valuable public datasets have been released that may enrich the training data of this model and thus improve its performance. In addition, the model shows limited robustness to click point location; when indicating different voxels in a lesion as the center voxel, the resulting lesion segmentations may differ.

In this study, we present ULS+, an improved version of the ULS model. Our contributions to the ULS model are twofold: (1) we extend the training data by incorporating additional publicly available whole-body CT lesion datasets, focusing specifically on lesions for which the original ULS model performed sub-optimally; and (2), we enhance robustness to click-point variation through train-time augmentation by sampling random lesion voxels as VOI center. We evaluate ULS+ against the ULS baseline in terms of both segmentation performance and click-point robustness, demonstrating its potential as a more reliable foundation for automated lesion analysis and longitudinal tracking.

\section{Materials and methods}
\begin{figure}[b]
    \centering
    \includegraphics[width=\linewidth]{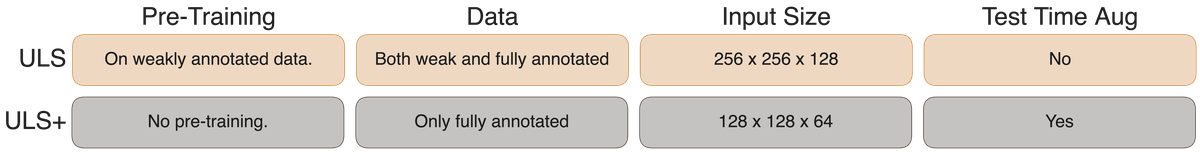}
    \caption{Differences between the ULS model and the ULS+ model.}
    \altText{Figure showing the difference between the ULS and and ULS+ model in terms of pre-training, data, input sizer and test-time augmentation.}
    \label{3886-fig:difs}
\end{figure}

The ULS+ model builds upon the original ULS baseline by modifying the training data, input configuration, and augmentation strategy to improve generalization and robustness to click-point variation. The main differences between the two models are visualized in Fig.~\ref{3886-fig:difs} and described in more detail below. 

\subsection{Datasets}
The original ULS model was trained on both a fully annotated and a weakly annotated dataset. To increase overall annotation quality and reduce training time, only the fully annotated part of this data was used for ULS+. In addition, ULS+ was trained on six more datasets. These datasets include both liver-specific sets and sets with multiple lesion types. Details on the added data can be found in Tab.~\ref{3886-tab:data}.

For each dataset, the center voxel of each lesion mask was determined, around which a VOI of $128 \times 128 \times 64$ voxels was cropped. If the cropped image included lesion masks unrelated to the central lesion, these masks were excluded. This resulted in one lesion mask per cropped image. 

\begin{table}[t]
  \caption{Overview of training data used for ULS+. *Only the fully annotated data of ULS was used. **Longitudinal-CT refers to task 2 of the AUTOPET challenge. ***Images from the liver, pancreas, colon and lung tasks were used.}
  \label{3886-tab:data}
  \begin{tabular*}{\textwidth}{l@{\extracolsep\fill}lll}
    \hline
    Dataset & Location & Number of lesions & Reference \\
    \hline
    ULS*        & Whole body        & 5737  & \cite{3886-DeGrauw2025} \\
    Longitudinal-CT**      & Whole body        & 4973  & \cite{3886-Gatidis2022} \\
    MSD***      & Whole body        & 1314  & \cite{3886-Antonelli2022} \\
    WORC GIST   & Gastrointestinal  & 248   & \cite{3886-Starmans2021} \\
    WORC CRLM   & Liver (metas.)    & 96    & \cite{3886-Starmans2021} \\
    CLM         & Liver (metas.)    & 479   & \cite{3886-Simpson2024} \\
    WAW-TACE    & Liver (prim.)     & 360   & \cite{3886-Bartnik2024} \\
    CECT        & Liver (prim.)     & 1274  & \cite{3886-Luo2025} \\
    MSWAL       & Abdominal         & 2638  & \cite{3886-Wu2026} \\
    \hline
  \end{tabular*}
\end{table}

\subsection{Lesion shifting}\label{3886-sec:lesion-shifting}
To increase robustness to click-point variability, we introduced a train-time augmentation strategy that simulates different user click points. For each lesion, we sampled two additional random click points within the lesion mask. We then cropped VOIs from the original CT scans centered around these new points, applying zero-padding if the crop extended beyond the original image boundaries. This yielded three spatially shifted variations of the same lesion for training.

\subsection{Training}
Training was performed on an NVIDIA L40S GPU using the nnUNet v2 framework~\cite{3886-Isensee2021} with a residual encoder (size L). We trained the model for 1000 epochs using standard settings, but with resampling disabled and the patch size set to equal the input image. Notably, we reduced the ULS+ input dimensions to $128 \times 128 \times 64$, compared to the $256 \times 256 \times 128$ used in the baseline. This reduction was prioritized to minimize inference latency for interactive clinical workflows on standard hardware, and because the larger volume proved redundant for click-centered tasks where only the central lesion is targeted.

Unlike the baseline, we omitted weakly supervised pretraining to simplify the pipeline, as prior experiments showed it yielded negligible improvements. Integrating other pretraining techniques was outside the scope of this study and remains a direction for future work.

\subsection{Evaluation}

To accurately compare the original ULS model to ULS+, we used the same test data as used in the ULS23 challenge. Additionally, we tested the models on a held-back part of the Longitudinal-CT dataset (20\%, split on patient level). While we recognize that part of the Longitudinal-CT set was used to train ULS+ and not to train the original ULS model, this dataset is still valuable because it contains more lesion types than the ULS23 challenge test set. This allows for more extensive organ-level evaluation. 

The original ULS model does not make use of test-time augmentation because this increases inference time. However, given its smaller input size, ULS+ allows for the use of test-time augmentation without rendering the model unusably slow in clinics. Therefore, this was turned on for the evaluation of ULS+.

To evaluate the robustness of the models to variations in user input, we simulated the variability of user-provided click-points using the mechanism explained in Section~\ref{3886-sec:lesion-shifting}. For each lesion, we generated three segmentations by cropping the input VOI around different points: one centered on the lesion's centroid ($P_{\text{normal}}$), and two centered on random points sampled within the lesion mask ($P_{\text{aug1}}$ and $P_{\text{aug2}}$) to simulate "off-center" clicks. This process results in slightly shifted input volumes for the model. The robustness score was defined as the mean pairwise Dice similarity between the resulting three segmentations:

\begin{equation}
\text{Robustness} = \frac{1}{3}\left[\text{Dice}(P_{\text{normal}}, P_{\text{aug1}}) + \text{Dice}(P_{\text{normal}}, P_{\text{aug2}}) + \text{Dice}(P_{\text{aug1}}, P_{\text{aug2}})\right]
\label{3886-equation}
\end{equation}

This metric ranges from 0 to 1, where higher scores indicate greater robustness to click-point placement and more consistent predictions despite shifts in the cropped input volume. 

\section{Results}

Table~\ref{3886-tab:results} summarizes performance on the ULS23 and Longitudinal-CT test sets. Overall, ULS+ consistently outperforms the original ULS model in terms of both Dice score and robustness across both datasets, with all improvements being statistically significant. The gains are more pronounced on the Longitudinal-CT dataset.

\begin{figure}[b]
    \centering
    \begin{subfigure}{\textwidth}
        \centering
        \includegraphics[width=\textwidth]{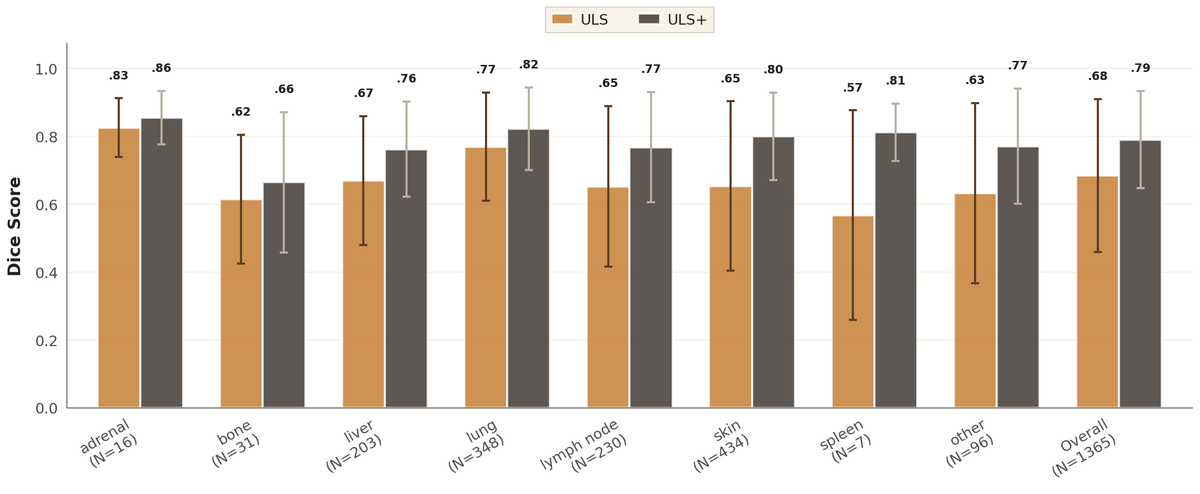}
        \caption{Longitudinal-CT test set.}
        \label{3886-fig:dice-Longitudinal-CT}
        \altText{Dice scores of original ULS and ULS+ model on Longitudinal-CT test dataset, by lesion location.}
    \end{subfigure}
        
    \begin{subfigure}{\textwidth}
        \centering
        \includegraphics[width=\textwidth]{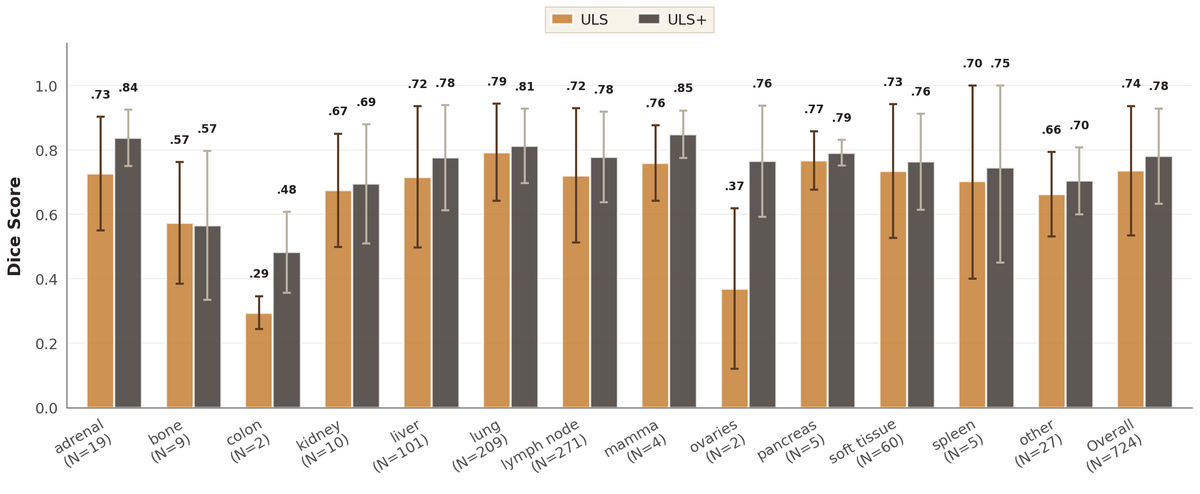}
        \caption{ULS23 Challenge test set.}
        \label{3886-fig:dice-uls23}
        \altText{Dice scores of original ULS and ULS+ model on ULS23 Challenge test dataset, by lesion location.}
    \end{subfigure}
    \caption{Dice scores of the original ULS model and ULS+ on two different test sets, stratified by lesion location.}
    \label{3886-fig:dice-comparison}
    \altText{Two bar charts comparing Dice scores between ULS and ULS+ models across anatomical lesion locations on Longitudinal-CT and ULS23 Challenge test sets.}
\end{figure}

\begin{figure}[b]
    \centering
    \begin{subfigure}{\textwidth}
        \centering
        \includegraphics[width=\textwidth]{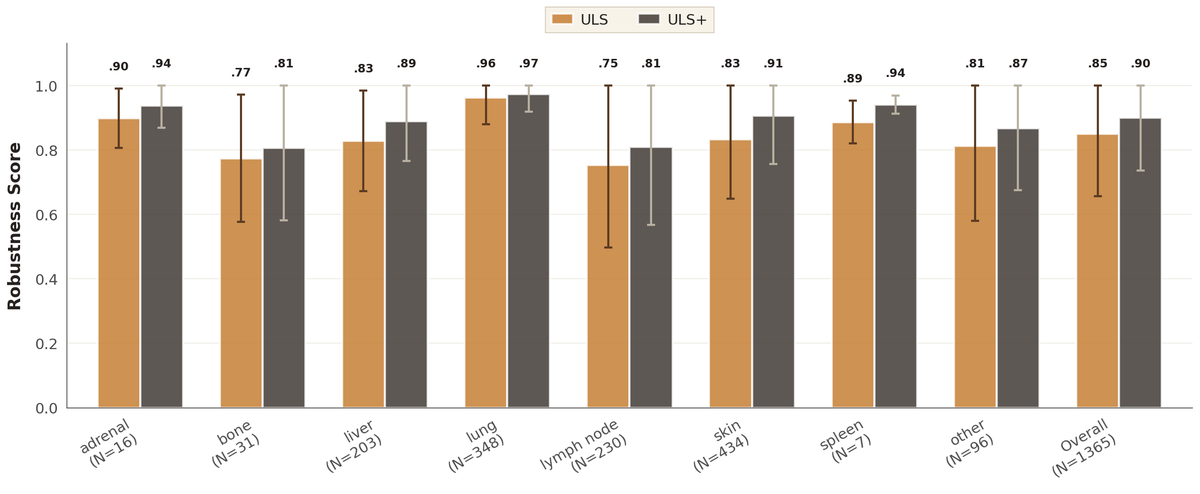}
        \caption{Longitudinal-CT test set.}
        \label{3886-fig:robustness-Longitudinal-CT}
        \altText{Robustness scores of original ULS and ULS+ model on Longitudinal-CT test dataset, by lesion location.}
    \end{subfigure}
        
    \begin{subfigure}{\textwidth}
        \centering
        \includegraphics[width=\textwidth]{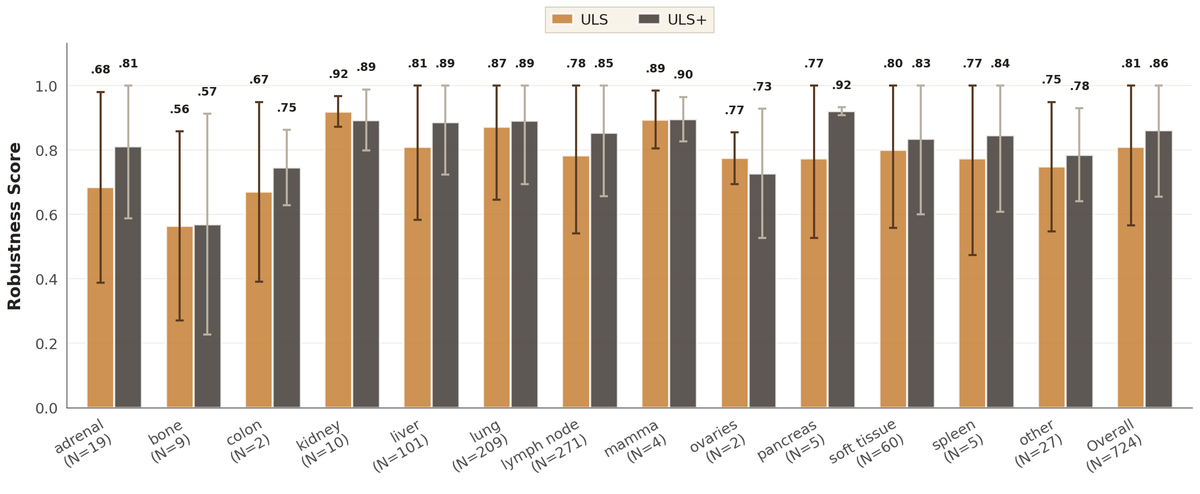}
        \caption{ULS23 Challenge test set.}
        \label{3886-fig:robustness-uls23}
        \altText{Robustness scores of original ULS and ULS+ model on ULS23 Challenge test dataset, by lesion location.}
    \end{subfigure}
    \caption{Robustness scores (as defined in Eq.~\ref{3886-equation}) of the original ULS model and ULS+ on two different test sets, stratified by lesion location.}
    \label{3886-fig:robustness-comparison}
    \altText{Two bar charts comparing robustness scores between ULS and ULS+ models across anatomical lesion locations on Longitudinal-CT and ULS23 Challenge test sets.}
\end{figure}

\begin{figure}[b]
    \centering
    \includegraphics[width=\linewidth]{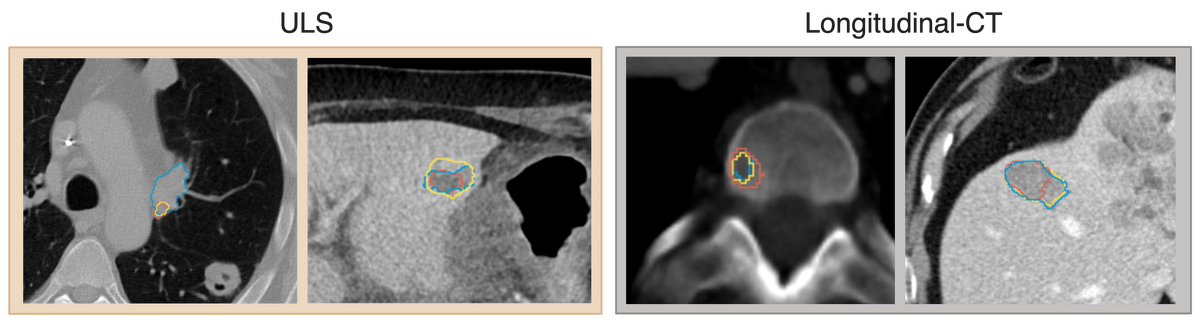}
    \caption{Example segmentations of the ULS and ULS+ model on both datasets on lesions from different locations. Blue: ground truth, red: ULS, yellow: ULS+. Note that for visibility of this figure, some lesions are zoomed in on while preserving surrounding context. Therefore, in this visualization, some lesions may not appear centered.}

    \label{3886-fig:visual}
    \altText{Grid of CT scan slices showing lesion segmentation comparisons with ground truth in blue, ULS predictions in red, and ULS+ predictions in yellow.}
\end{figure}

Per-lesion plots in Fig.~\ref{3886-fig:dice-comparison} (Dice) and Fig.~\ref{3886-fig:robustness-comparison} (robustness) show that ULS+ generally shifts scores toward higher dice and robustness values, but it is not uniformly better for every lesion type (e.g., adrenal and bone lesions). These exceptions occur in lesion categories with very small sample counts. Qualitative examples of both successful and challenging cases are shown in Fig.~\ref{3886-fig:visual}.

ULS+ was submitted to the test phase of the ULS23 challenge and achieved first place on the leaderboard with a challenge score of 0.749. For further details on the challenge, we refer the reader to the original ULS paper~\cite{3886-DeGrauw2025}

\begin{table}[t]
  \centering
  \caption{Comparison of the original ULS model and ULS+ on both the ULS23 and Longitudinal-CT datasets. *Indicates statistical significance ($p < 0.0001$ using a paired two-tailed t-test with Bonferroni correction for multiple testing).}
  \label{3886-tab:results}
  \begin{tabular*}{\textwidth}{l @{\extracolsep{\fill}} c c c c}
    \hline
    & \multicolumn{2}{c}{ULS23} & \multicolumn{2}{c}{Longitudinal-CT} \\
    \cline{2-3} \cline{4-5}
    & Dice & Robustness & Dice & Robustness \\
    \hline
    ULS baseline & $0.74 \pm 0.20$ & $0.81 \pm 0.24$ & $0.68 \pm 0.23$ & $0.85 \pm 0.19$ \\
    ULS+         & $0.78 \pm 0.15$\rlap{*} & $0.86 \pm 0.20$\rlap{*} & $0.79 \pm 0.14$\rlap{*} & $0.90 \pm 0.16$\rlap{*} \\
    \hline
  \end{tabular*}
\end{table}

\section{Discussion}

This study introduces ULS+, an improved version of the Universal Lesion Segmentation (ULS) model, designed to enhance segmentation accuracy and robustness to click-point variation. Despite using a smaller input size and no pretraining, ULS+ achieved higher Dice scores and improved robustness to click point location on two separate datasets. Improvements were relatively homogeneous across lesion location, which may be attributed to the fact that the newly added training data contains cases of lesions across the whole body, providing enrichment for all these lesion types. 

The ULS+ framework incorporates four primary updates: the inclusion of expanded, fully annotated multi-organ data, the implementation of click-point augmentation, the transition to a $128 \times 128 \times 64$ input size and test-time augmentation. While these changes collectively coincide with the observed performance gains, the specific contribution of each factor has not yet been isolated. Consequently, a systematic ablation study and a dedicated same-dimension comparison are planned as future work to quantify the individual impact of these components.

Although several of the newly added datasets were specific to liver lesions, the observed performance gains were not limited to this organ. The similar improvement in Dice score across all lesion types suggests that the benefits of the new data extend beyond organ-specific information, strengthening the model’s capacity for whole-body lesion segmentation. The slightly lower click-point robustness in skeletal and undefined lesions may reflect the intrinsic challenges of these structures, such as small size and unclear lesion boundaries. 

It should be noted that in addition to better performance, the ULS+ model is faster than the original. ULS takes input images of $256 \times 256 \times 128$, whereas ULS+ uses an input image size of $128 \times 128 \times 64$. Inference with ULS+ takes around 0.5 seconds on an NVIDIA A100 SXM4 GPU when applying test-time augmentation. This combination of increased performance and high speed makes it more realistically applicable in clinical settings, where time efficiency is highly valued. The smaller input size introduces the inherent limitation that lesions extending beyond the input field of view cannot be fully segmented in a single forward pass. A sliding-window inference strategy would address this, but was outside the scope of this study. 

Our study shows the value of iterative retraining and updating AI models with new, relevant data. As more and more valuable, well-annotated medical imaging data becomes publicly available, preexisting models may benefit from this added information without requiring significant adaptations to architecture. This paradigm supports the sustainability and continual improvement of clinical AI tools, ensuring that deployed systems evolve alongside advances in data availability and imaging practice. By maintaining a cycle of data-driven updates and clinical validation, ULS+ can serve as a foundation for robust, transparent, and clinically relevant lesion segmentation models.

To facilitate reproducibility the code, data and trained model weights are available at \url{https://github.com/DIAGNijmegen/oncology-uls-plus}.

\printbibliography

\end{document}